\documentclass[runningheads]{llncs}
\usepackage{comment}
\usepackage{bbding}
\usepackage{enumerate} 
\usepackage{graphicx}

\usepackage{hyperref}

\begin{document}
\title{Attention to Refine through Multi-Scales for Semantic Segmentation}
\titlerunning{Attention to Refine}
%
\author{Shiqi Yang \and Gang Peng\Envelope}

\authorrunning{S. Yang and G. Peng}
%
\institute{Key Laboratory of Ministry of Education for Image Processing and Intelligence Control, School of Automation,\\Huazhong University of Science and Technology, Wuhan, China 
\email{albert\_yang@hust.edu.cn, penggang@hust.edu.cn}
}
\maketitle              
\begin{abstract}
This paper proposes a novel attention model for semantic segmentation, which aggregates multi-scale and context features to refine prediction. Specifically, the skeleton convolutional neural network framework takes in multiple different scales inputs, by which means the CNN can get representations in different scales. The proposed attention model will handle the features from different scale streams respectively and integrate them. Then location attention branch of the model learns to softly weight the multi-scale features at each pixel location. Moreover, we add an recalibrating branch, parallel to where location attention comes out, to recalibrate the score map per class. We achieve quite competitive results on PASCAL VOC 2012 and ADE20K datasets, which surpass baseline and related works.

\end{abstract}
\section{Introduction}
With the booming of deep learning, many visual tasks have made significant progress. For instance, semantic segmentation, also known as image labeling or scene parsing which aims at giving label for each pixel, has made great breakthroughs in recent years. Efficient semantic segmentation can facilitate plenty of other missions such as image editing.

Recent approaches for semantic segmentation are all almost based on Fully Convolutional Network (FCN)~\cite{long2015fully}, which outperforms the traditional methods by replacing the fully connected layers with convolutional layers in classification network. The follow-up works have extended the FCN from several points of view. Some works~\cite{badrinarayanan2017segnet,noh2015learning} have introduced the coarse-to-fine structure with upsample modules like deconvolution to give the final mask prediction. And due to the usage of pooling layer, spatial size has decreased largely, for which dilated (or atrous) convolution~\cite{yu2015multi,chen2018deeplab} has been employed to increase the resolution of intermediate features and hold the same receptive field simultaneously.

Other works mainly focus on two directions. One is to post-process the prediction from the CNN through Conditional Random Field (CRF) to get smooth output. These works~\cite{arnab2016higher,chen2018deeplab,zheng2015conditional} are actually ameliorating the localizing ability of the framework. Another direction is to ensemble multi-scale features. Because features from lower layers in CNN have more spatial information and ones from deeper layers have more semantic meaning and less location information, it is rational to integrate representations from various positions since location information is important for semantic segmentation. The first type method for multi-scale combines features from different stages with skip connection to get fused features for mask prediction, such as~\cite{chen2018deeplab,long2015fully,xia2016zoom}. And another type is to resize input to several scales and pass each one with a shared network, it will produce final prediction using the fusion of multi-stream resulting features. There are also methods trying to exploit the capability of global context information, like ParseNet~\cite{liu2015parsenet} which adds a global pooling branch to extract contextual features. And PSPNet~\cite{zhao2017pyramid} adopts a pyramid pooling module to embed global context information to achieve accurate scene perception.

Attention model has been all the rage in natural language processing area, such as~\cite{bahdanau2014neural}, and it has also shown its effectiveness in computer vision and multimedia community recently~\cite{xu2015show,wang2017residual,chen2017attentive,song2018neural}. It allows model to focus on specific relevant features. Attention-to-scale~\cite{chen2016attention} is the first approach to introduce attention model into semantic segmentation for multi-scale. It takes in different scale inputs. For each scale, the attention model produces a weight map to weight features at each location, and the weighted sum of score maps across all scales is then used for mask prediction. But it only utilizes the feature from specific layer to generate attention, which may omit many contextual details, and this can not ensure that the attention model can guide network to get precise results.

Referring to attention-to-scale, we propose a new attention model in this paper, which aslo takes in multi-scale inputs but integrates features from different layers, similar to hypercolumns~\cite{hariharan2015hypercolumns}. The attention model has two branch outputs, \emph{i.e.}, one for location attention through which it drives network to focus on large objects or regions for small scale input and pay attention to small targets for large scale just like attention-to-scale, another branch is to recalibrate the score map per class since resulting features from several stages carry contextual information. The outputs from attention model will be applied to multi-scale stream predictions, and final mask prediction is a weighted sum of all these streams.

Our contributions are two aspects as follows:
\begin{enumerate}[(1)]  
	\item We introduce a novel attention model into multi-scale streams semantic segmentation framework, the final mask prediction is produced by merging the predictions from multiple streams.
	\item The attention model utilizes fused features from different positions of CNN, which carry more contextual information, and has two branch outputs, where one is for location attention and another is for recalibrating.
\end{enumerate}

\section{Proposed Methods}
\subsection{Attention Model with Multi-Scales}
\label{sec2.1}

\begin{figure}
	\includegraphics[width=1\textwidth]{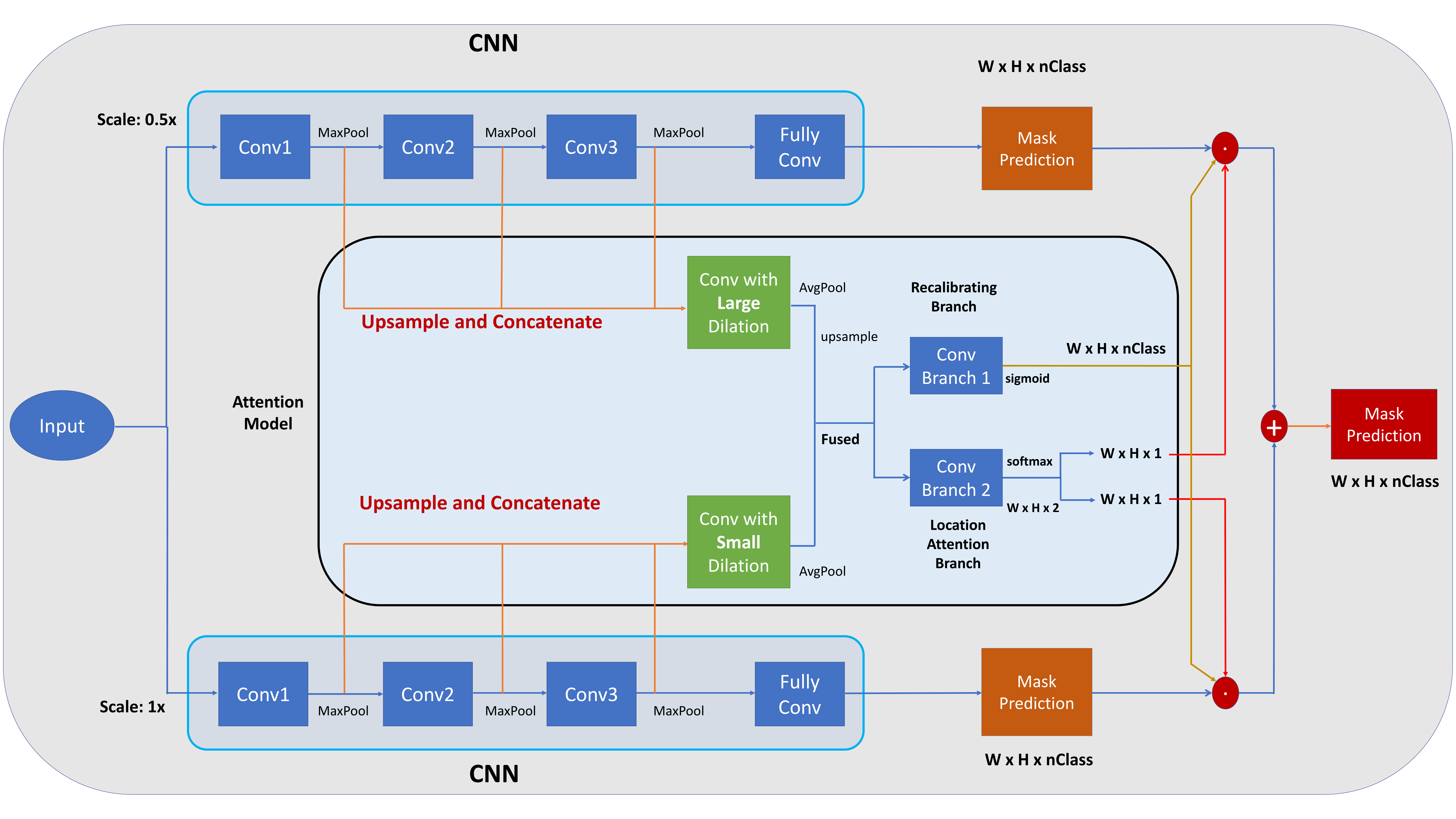}
	\caption{Architecture of semantic segmentation framework with the proposed attention model. The attention model takes in features from different stages in CNN just like hypercolumns~\cite{hariharan2015hypercolumns}, and then it adopts convolutional layer with different dilation to process features for each scale respectively. Attention model produces two kinds of weight maps which are applied to multiple streams predictions. The final mask prediction is a sum of all streams.} \label{fig1}
\end{figure}

Like we mention before higher-layer features contain more semantic information and lower ones carry more location information. Fusion of information from several spatial scales will improve the accuracy of prediction in semantic segmentation. In addition, multi-scale aggregation also catch more contextual representations since some operations like pooling will dispose of the global context information, leading to local ambiguities which will be discussed later. It is the reason why multi-scale fusion gained a lot of popularity.

Since our work is extended from attention-to-scale~\cite{chen2016attention}, here we give a brief review on it. In attention-to-scale, the images are resized to several scales which will be fed to a weight-shared CNN, and the attention model takes as input the directly concatenating features from penultimate layer in each scale stream. The attention model consists of two convolutional layers and will produce \emph{n} channels scores map, where \emph{n} means the number of input scales. The attention model is expected to adaptively find the best weights on scales. But there exists some problems. The features from penultimate layer surely contain semantic representations, but they lack essential localization and global information fed to the attention model to achieve precise prediction. And we also posit that simply concatenating features from certain position is not conducive to lead the attention model to learn soft weight across scales. Seeing that the attention model is to put large weights on the large object or region in small-scale stream and gives large weights to the small targets in large-scale stream, we think it is rational to handle features from different scales respectively before integrating them.

\begin{figure}
	\includegraphics[width=1\textwidth]{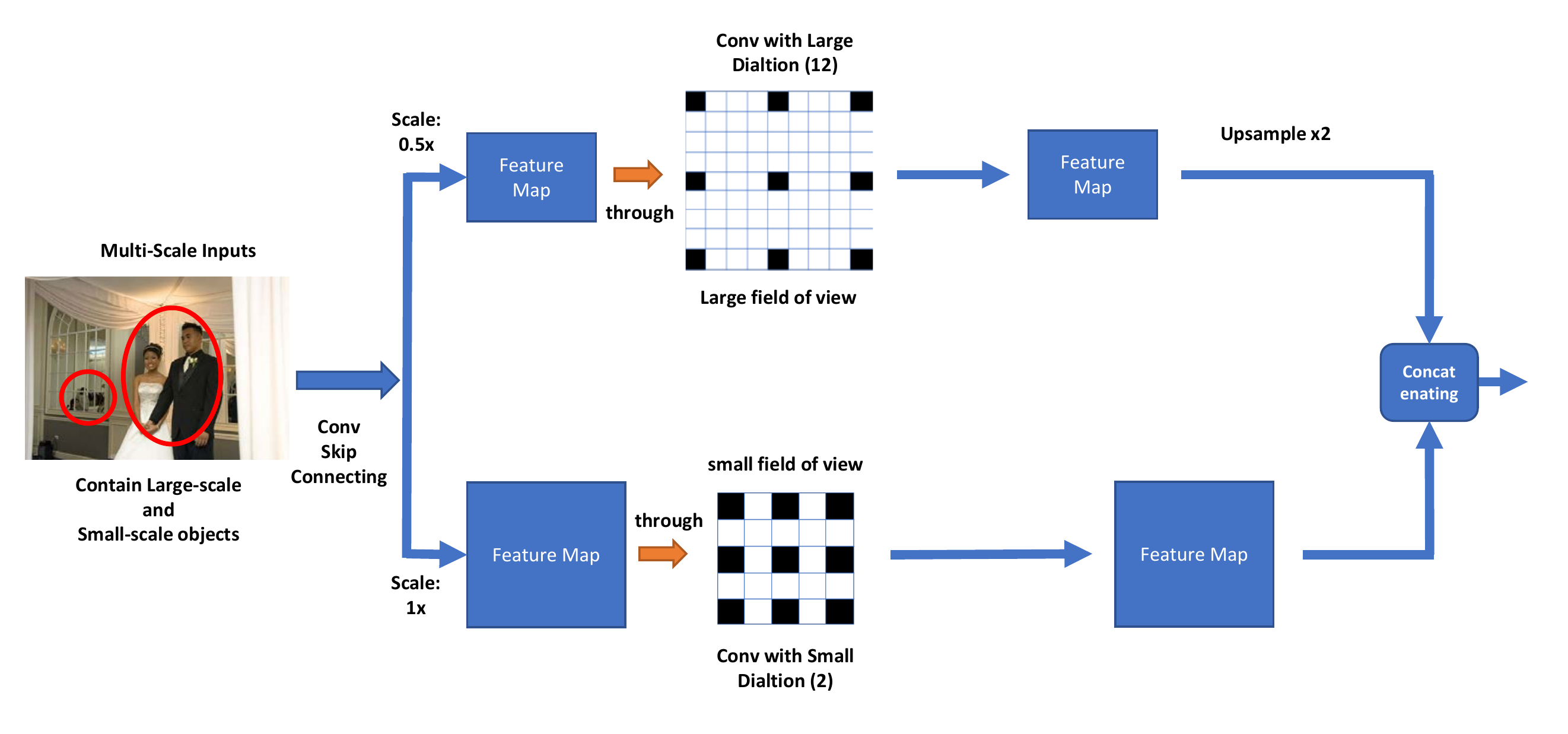}
	\caption{Convolution with different dilation for different scale. Convolution with large dilation has large field of view while convolution with small dilation has small field of view.} \label{fig2}
\end{figure}

Inspired by hypercolumns, we adopt the the philosophy of it. Like depicted in Figure~\ref{fig1}, features from different stages in CNN get upsampled to same size and then we concatenate them all. To keep computation cost at bay, we choose the size of features after two pooling operation as the appointed resolution to do upsampling by bilinear interpolation. Through this way, the acquired features carry more localization and context information. 

It is well-known that the structure of network has an impact on the range of pixels of the input image which correspond to a pixel of the feature map. In other words, filters will implicitly learn to detect features at specific scales due to the fixed receptive field. To accomplish our motivation of attention model which is to adaptively put weights on corresponding scale, we add a unique convolutional layer with unequal dilation for each scale. This process is demonstrated in Figure~\ref{fig2}. Convolution with large dilation has large field of view (FOV) and is expected to catch the long-span interlink of pixels for large scale object or region in small scale stream, and small dilation convolution is deployed to encode target of small scale in large scale stream. After the dilated convolution, the features will be concatenated, resulting one contains much more abundant and context information.

By the way, the two-stream CNNs in Figure~\ref{fig1} are actually the same one when implemented in practice, just like Siamese Network.

\subsection{Two branch outputs of Attention Model}
The concatenated features will go through two parallel convolutional branches: location attention branch and recalibrating branch.

In common with attention-to-scale, the attention model will produce soft weights for multiple scales (we refer to it as location attention). Assuming the number of input scale is \emph{n}, and the size of mask prediction, which is denoted as $P^s$ for scale \emph{s}, is W x H, \emph{nClass} means the class number of the objects. The location attention output by the model is shared across all channels. After the refinement of local attention, the mask predictions, denoted as $M^s_i$, are described as:

\begin{equation}
M^s_{i,c} = \sum^n_{s=1} l^s_i \cdot P^s_{i,c}
\label{eq1}
\end{equation}

The $l^s_i$ is computed by:
\begin{equation}
l^s_i = \frac{\exp{(wl_i^s)}}{\sum^n_{j=1}{\exp{(wl_i^j)}}}
\label{eq2}
\end{equation}
where $wl_i^s$ is the score map produced by the location attention branch at position $i \in [0, W * H-1]$ for scale \emph{s}, before the softmax layer of course.

And since the fused features fed to the attention model contain context information, we want to make full use of them to eliminate some degrees of class ambiguity, \emph{i.e.}, to utilize contextual relationship to enhance the ability of classification. The lack of ability to collect contextual information may increase the chance of misclassification in certain circumstances. To take an example, neural network sometimes tends to take apart a large-scale object into several regions of different classes~\cite{li2017foveanet}, or maybe classify a boat on the river as a car and so on in scene parsing~\cite{zhao2017pyramid} (these can be observed among visualization results in section~\ref{sec3.1}). To deal with these issues, we add a recalibrating branch parallel to location attention. It has the same architecture as location attention branch which means containing two convolutional layers, except that output channel changes to \emph{nClass} and \emph{sigmoid} activation is deployed instead of softmax. This branch aims to find the interdependencies between adjacent objects or regions using the integrating features, and its output is used for recalibrating the score maps before the location attention refinement. Because the contextual relationship stay the same in different scale, the recalibrating outputs are shared across all scales. So the final mask prediction for each stream can be described as:

\begin{equation}
M^s_{i,c} = \sum^n_{s=1} l^s_i \cdot [P^s_{i,c} \otimes wr_{i,c}]
\label{eq3}
\end{equation}
where the $\otimes$ means element-wise multiplication and $wr_{i,c}$ means output in position \emph{i} in channel $c \in [0, n-1]$ produced by recalibrating branch. Another choice for recalibrating branch is to predict bias per position in each channel instead of multiplication. But it will bring around 1\% performance decrease according to our experiment.

And the ultimate mask prediction is as below, where $M^s$ is the mask prediction of scale \emph{s}:

\begin{equation}
M_{final} = \sum^n_{s=1} M^s
\label{eq4}
\end{equation}

As for the loss function, we follow the setting of attention-to-scale, \emph{i.e.}, the total loss function is sum of 1+\emph{S} cross entropy loss functions for segmentation, where \emph{S} symbolizes number of scales and one for final prediction.

\section{Experimental Results}
We experiment our method on two benchmark datasets: PASCAL VOC 2012~\cite{pascal-voc-2012} and ImageNet scene parsing challenge 2016 dataset~\cite{zhou2016semantic} (it is from ADE20K~\cite{zhou2017scene}, hereinafter we refer to it as ADE20k). 

For all training, we only train the network with 2 scales, \emph{i.e.}, 1x upsample and 0.5x upsample. As for the different dilation, we set it to 2 for small scale and 12 for large scale. And we use the poly learning rate policy~\cite{liu2015parsenet}, meaning current learning rate is computed by multiplying $(1 - \frac{iter}{max\_iter})^{power}$ to base learning rate, where the power is set to 0.9. We refer to the layers in the last stage where gives mask prediction as decoder, layers previous to decoder are encoder. Learning rate of decoder is 10 times that of encoder. All experiments are implemented using PyTorch on a NVIDIA TITAN Xp GPU.

\subsection{PASCAL VOC 2012}
\label{sec3.1}
\setlength{\tabcolsep}{14pt}
\begin{table}
	\begin{center}
		\caption{Results on PASCAL VOC 2012 validation set. There exists 2 scale streams: 1x and 0.5x. The mIoU means mean intersection of union~\cite{long2015fully}.}\label{tab1}
		\begin{tabular}{|l|l|}
			\hline
			Method &  mIoU \\
			\hline
			\hline
			Baseline (DeepLab-LargeFOV)  & 61.40\%\\
			\hline
			Merged with MaxPooling  & 63.88\%\\
			\hline
			Merged with AvgPooling  & 64.07\%\\
			\hline
			Attention-to-Scale & 64.74\%\\
			\hline
			\textbf{Our method}	& \textbf{67.98\%}\\
			\hline
		\end{tabular}
	\end{center}
\end{table}

The PASCAL VOC 2012~\cite{pascal-voc-2012} segmentation dataset consists of 20 foreground object classes and a background class. The PASCAL VOC 2012 dataset we use is augmented with extra annotation by Hariharan \emph{et al}.~\cite{hariharan2011semantic}, resulting in 10582 training images. In experiment we report performance results on original PASCAL VOC 2012 validation set.

DeepLab-LargeFOV~\cite{chen2014semantic} is chosen as base model. Since our work is extended from attention-to-scale, in order to compare fairly, we reproduce the DeepLab-LargeFOV and attention-to-scale based on it by ourselves, following the set of attention-to-scale~\cite{chen2016attention}. All these experiments use VGG16~\cite{simonyan2014very} as skeleton CNN, which is pretrained on ImageNet. Our reproduction of them yields performance of 61.40\% and 64.74\% on the validation set respectively. The performance of attention-to-scale is lower than original paper, but the follow-up experiments still can verify effectiveness of our proposed method since ours is directly built on attention-to-scale. Noted that both of attention-to-scale and our work adopt extra supervision, meaning adding softmax loss function for each scale stream. The results of experiment are demonstrated in Table~\ref{tab1}.

\begin{table}
	\begin{center}
		\caption{Ablation study for proposed method on PASCAL VOC 2012. The multi-stage means hypercolumns-like feature integration from different positions. Diverse dilation means utilizing different dilated convolution for multi-scale features. Extra branch means adding recalibrating branch. *-The base model is actually attention-to-scale. $\dagger$-No diverse dilations means using standard convolution instead.}\label{tab2}
		\setlength{\tabcolsep}{4mm}\resizebox{\textwidth}{10mm}{
		\begin{tabular}{|l|l|l|l|l|l|}
			\hline
			Method & multi-stage & diverse dilations$\dagger$ & location attention & extra branch & mIoU\\
			\hline
			\hline
			Base model*  & & & $\surd$ & & 64.74\%\\
			\hline
			Base model+  &$\surd$& & $\surd$ & & 65.80\%\\
			\hline
			Base model++  & $\surd$ & $\surd$ & $\surd$ & & 66.83\%\\
			\hline
			\textbf{Our method} & $\surd$ & $\surd$ & $\surd$ & $\surd$ & \textbf{67.98\%}\\
			\hline
		\end{tabular}}
	\end{center}
\end{table}
\begin{figure}
	\begin{center}
		\includegraphics[width=1\linewidth]{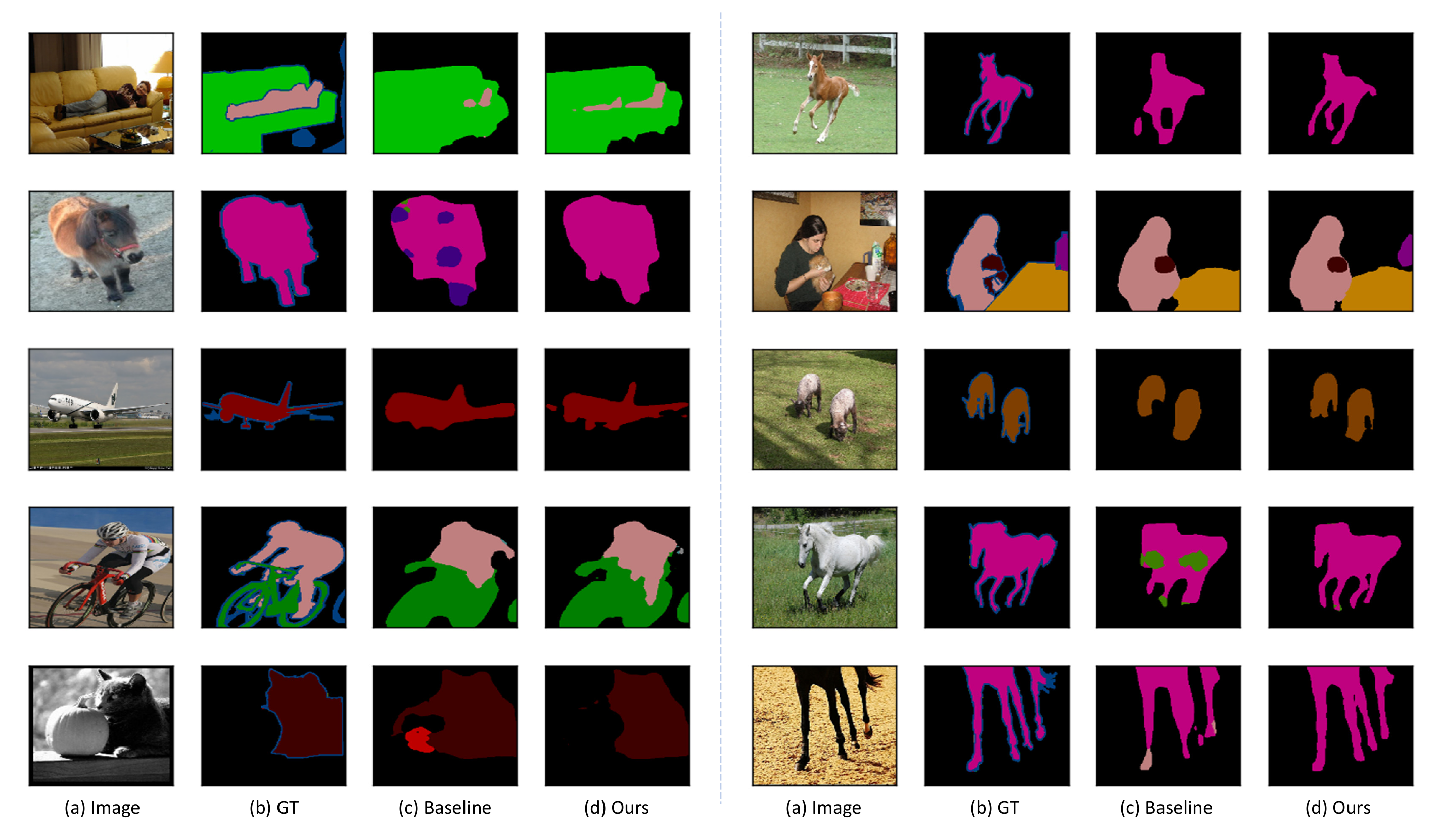}
	\end{center}
	\caption{Representative visual segmentation results on PASCAL VOC 2012 dataset. Images are from train and val set. GT means ground truth, and baseline means attention-to-scale approach. Our proposed method produces more accurate and detailed results.}
	\label{fig3}
\end{figure}

Merged with Pooling in Table~\ref{tab1} means adopting pooling operation as fusion approach for multi-scale stream instead of attention model. It can be seen that our method surpasses baseline and attention-to-scale by 6.58\% and 3.24\% respectively. Furthermore, we conduct additional experiments for ablation study of each module in our method. We cut off certain modules from our proposed method, re-train and report the performance of remainder, which is shown in Table~\ref{tab2}. Please noted that base model without all these modules is actually attention-to-scale approach. As you can see, the modules we design indeed take effect on segmentation task.

Since the attention-to-scale has verified the motivation which we share with by visualizing weight maps produced by the attention model, we don't replicate this experiment on our proposed model. Turning to qualitative results, some representative visual comparisons are provided between attention-to-scale and our method in Figure~\ref{fig3}. We observe that unlike attention-to-scale, our method can get finer contour in some cases and probability of breaking down a large-scale object into several pieces decreases. Our results contain much more detailed structure and more accurate pixel-level categorization, which we posit it comes from the utilization of multi-scale and context information as well as the extra branch.

\subsection{ADE20K}
ADE20K dataset first shows up in ImageNet scene parsing challenge 2016. It is much more challenging since it has 150 labeled classes for both objects and background scene parsing. It contains around 20K and 2K images in the training and validation sets respectively. 

\begin{table}
	\begin{center}
		\caption{Results on ADE20K validation set. *- Two multi-scale attention methods take as input two scale streams: 1x and 0.5x.}\label{tab3}
		\setlength{\tabcolsep}{6mm}\resizebox{\textwidth}{12mm}{
		\begin{tabular}{|l|l|l|}
			\hline
			Method &  mIoU & Pixel Accuracy\\
			\hline
			\hline
			ResNet34-dilated8 (Baseline)  & 32.67\%  &  76.41\%\\
			\hline
			Baseline + attention-to-scale*  & 35.11\%  &  76.82\%\\
			\hline
			Baseline + PSP  & 36.43\%  &  78.01\%\\
			\hline
			\textbf{Baseline + our attention model*} & \textbf{37.07\%}  &  78.57\% \\
			\hline
			\textbf{Baseline + our attention model* + PSP} & \textbf{38.21\%}  &  79.29\% \\
			\hline
		\end{tabular}}
	\end{center}
\end{table}

We deploy ResNet34-dilated8~\cite{yu2015multi} (not resnet50 because of limited GPU memory) as base CNN to investigate several different methods. Besides applying attention-to-scale and our proposed attention model, we also experiment on Pyramid Scene Parsing (PSP)~\cite{zhao2017pyramid} module as a comparison, which is a state-of-the-art approach on ADE20K dataset to the best of our knowledge. The experiment results are presented in Table~\ref{tab3}. The PSP here doesn't contain auxiliary loss in original paper. We can see that our proposed attention model outperforms other methods, and achieves 4.40\% improvement on mIoU over baseline. Besides, we also embed both the PSP module and proposed attention module in baseline and it obtains further performance improvement.

\section{Conclusion}
In this paper, we propose a novel attention model for semantic segmentation. The whole CNN framework takes in multi-scale streams as input. Features from different stage of CNN are fused, then resulting one in each scale goes through convlutional layers with different dilation, which are expected to catch distinctive context relationship for different scales. After that, all these features get concatenated and resulting one is fed into two parallel convolution output branches of the attention model. One of the branches is location attention, aiming to pay soft attention to each location across channels. Another one is designed to fully utilize contextual information to deal with class ambiguity by recalibrating the prediction per location for each class. Experiments on PASCAL VOC 2012 and ADE20K show that proposed method make a significant improvement.

%
%
%
%
\bibliographystyle{splncs04}
\bibliography{att}
\end{document}